\documentclass[journal,final]{IEEEtran}
\ifCLASSINFOpdf
 \usepackage[pdftex]{graphicx}
\else
\fi
\usepackage{amsmath}
\usepackage{amsfonts}

\usepackage{booktabs}

\begin{document}
%
\title{SpatialFlow: Bridging All Tasks for Panoptic Segmentation}
%
%
%

\author{Qiang Chen,
        Anda Cheng,
        Xiangyu He,
        Peisong Wang,
        and Jian Cheng
\thanks{Qiang Chen, Anda Cheng, Xiangyu He, Peisong Wang,  and Jian Cheng are with the National
Laboratory of Pattern Recognition (NLPR), Institute of Automation Chinese Academy
of Sciences (CASIA) and School of Artificial Intelligence, University of Chinese Academy of Sciences (UCAS), Beijing, China. (e-mail: qiang.chen@nlpr.ia.ac.cn; chenganda2017@ia.ac.cn; xiangyu.he@nlpr.ia.ac.cn; peisong.wang@nlpr.ia.ac.cn; 
jcheng@nlpr.ia.ac.cn).}
\thanks{Corresponding author: Jian Cheng (jcheng@nlpr.ia.ac.cn)}
\thanks{Copyright \text{\textcopyright} 20xx IEEE. Personal use of this material is permitted. However, permission to use this material for any other purposes must be obtained from the IEEE by sending an email to pubs-permissions@ieee.org.}}

%
%

\markboth{JOURNAL OF IEEE TRANSACTIONS ON CIRCUITS AND SYSTEMS FOR VIDEO TECHNOLOGY}%
{}
%



\maketitle

\begin{abstract}
Object location is fundamental to panoptic segmentation as it is related to all things and stuff in the image scene. Knowing the locations of objects in the image provides clues for segmenting and helps the network better understand the scene. How to integrate object location in both thing and stuff segmentation is a crucial problem. In this paper, we propose spatial information flows to achieve this objective. The flows can bridge all sub-tasks in panoptic segmentation by delivering the object's spatial context from the box regression task to others. More importantly, we design four parallel sub-networks to get a preferable adaptation of object spatial information in sub-tasks. Upon the sub-networks and the flows, we present a location-aware and unified framework for panoptic segmentation, denoted as SpatialFlow. We perform a detailed ablation study on each component and conduct extensive experiments to prove the effectiveness of SpatialFlow. Furthermore, we achieve state-of-the-art results, which are $47.9$ PQ and $62.5$ PQ respectively on MS-COCO and Cityscapes panoptic benchmarks. Code will be available at https://github.com/chensnathan/SpatialFlow.\end{abstract}


%
\IEEEpeerreviewmaketitle

%
%
%
%

\section{Introduction} \label{intro}

\IEEEPARstart{R}{eal-world} vision systems, such as autonomous driving or augmented reality, require a rich and complete understanding of the image scene. However, neither detect and segment the objects in the image nor segment the image semantically can provide a global view of the image scene. Considering the tasks as a whole is a step forward to real-world vision systems. In the pre-deep learning era, there are classical vision tasks, such as scene understanding~\cite{detectorcrfs,wholescene}, considering object detection and semantic segmentation jointly. With the development of deep learning, instance and semantic segmentation have been widely studied and improved, while studies of the joint task have been left behind. Recently, ~\cite{panopticsegmentation} proposed the panoptic segmentation task to unify two segmentation tasks. In this task, countable objects such as persons, animals, and tools are considered as {\em things}, while amorphous regions of similar texture or material such as grass, sky, and road are referred to as {\em stuff}. It draws the attention of the vision community and pushes the deep vision systems a step forward towards applications in the real-world scenarios.  
\begin{figure}
  \centering
  \includegraphics[width=0.5\textwidth]{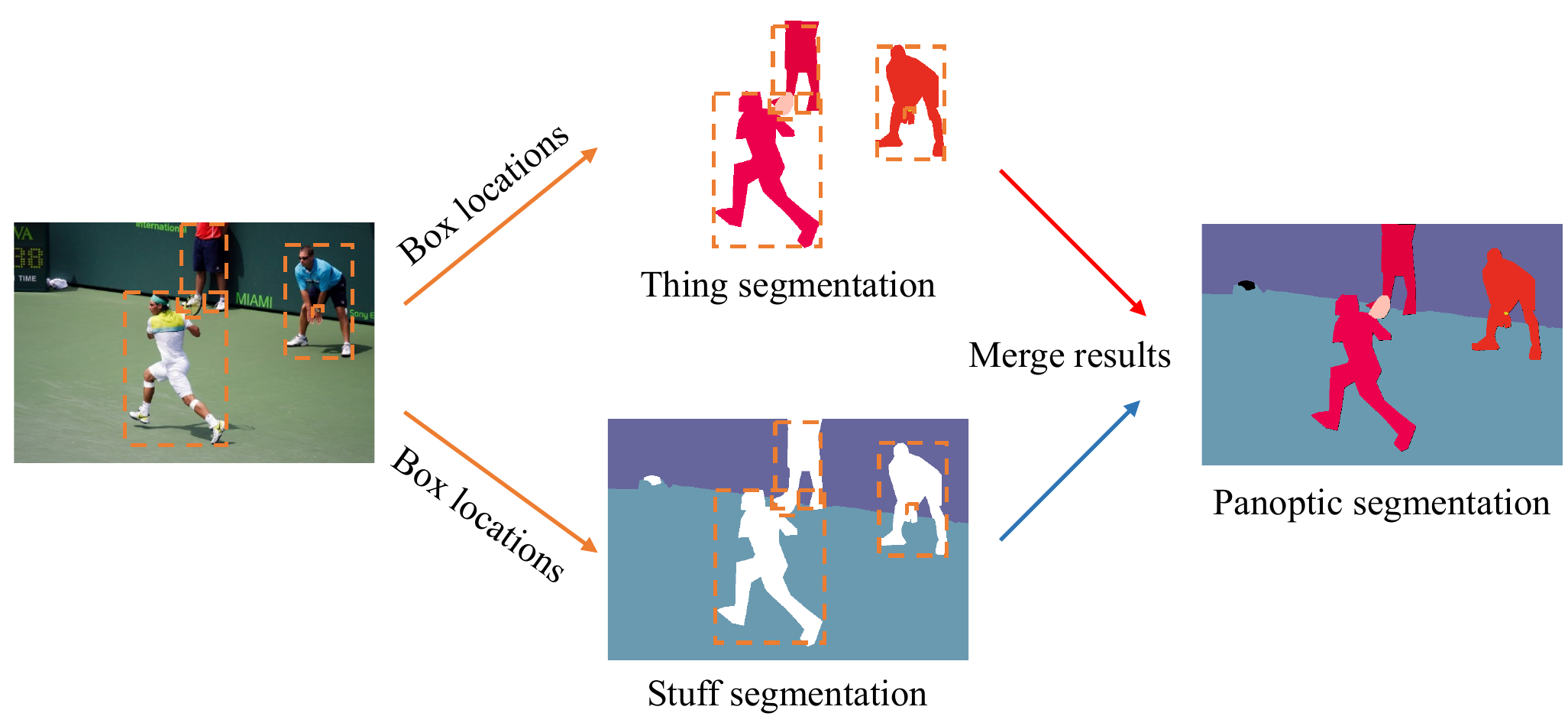}
  \caption{An illustration of the panoptic segmentation task. We also provide the bouding box for each object in the image and add process to integrate box location to both thing and stuff segmentation.}
   \label{fig0}
\end{figure}

Panoptic segmentation aims to assign all pixels in an image with a semantic and an instance id, which is a challenging task as it requires a global view of segmentation. In ~\cite{panopticsegmentation}, the authors tried to solve the task by adopting two independent models, Mask R-CNN~\cite{maskrcnn} and PSPNet~\cite{pspnet}, for thing and stuff segmentation\footnote{Refer to as instance and semantic segmentation, in this paper, we use the thing and the stuff to emphasize the tasks in panoptic segmentation.} respectively. Then, they applied a heuristic post-processing method to merge the segmentation outputs of two tasks, as illustrated on the right side of Figure~\ref{fig0}. These two independent models ignore the underlying relationship between things and stuff and bring computation budder into the framework. Recently, several works~\cite{jsisnet,tascnet,panopticfpn,aunet,upsnet,panopticranking} follow~\cite{panopticsegmentation} and try to build a unified pipeline for panoptic segmentation via sharing the backbone between two segmentation tasks. 

However, most of the recent works focus on how to combine the outputs of segmentation tasks properly, failing to highlight the significance of object location when training networks. As demonstrated in the literature, the spatial information of objects can boost the performance of algorithms in object detection~\cite{enrichedssd,garpn}, instance segmentation~\cite{htc,yolact}, and semantic segmentation~\cite{boxsup,boxdriven}. {\em Our key insight} is that, as a combination of these tasks, panoptic segmentation can benefit from delivering spatial information of objects among its sub-tasks. We illustrate the process of performing panoptic segmentation with box locations in Figure~\ref{fig0}. 

A crucial question then arises: how to integrate spatial information into the segmentation tasks seamlessly? To fulfill this goal, we propose to combine object location by explicitly delivering the spatial context from the box regression task to others. Based on this, we introduce a new unified framework for panoptic segmentation by fully leveraging the reciprocal relationship among detection, thing segmentation, and stuff segmentation. Two primary principles are considered as follows.

{\em First, keep the spatial context in pixel-level before segmenting things and stuff}. Although thing and stuff segmentation can complement one another, the format of dominated features in these two segmentation tasks may be inconsistent. The instance-level features control the thing segmentation, while the pixel-level features guide the stuff segmentation. The instance-level spatial context may not be suitable for stuff segmentation, given the format of its dominant feature. Besides, instances can be overlapping, which makes it hard to map them back to pixel-level. Based on this principle, we resort to one-stage detector RetinaNet~\cite{retinanet} instead of two-stage detector Faster R-CNN~\cite{fasterrcnn}. It prevents the spatial context of objects from being in the format of instance-level before performing segmentation tasks. Then, we extend RetinaNet with task-specific heads - thing head~\cite{retinamask} and stuff head~\cite{panopticfpn} to perform thing and stuff segmentation. In task-specific heads, the spatial context can be instance-level for things and be pixel-level for stuff.

{\em Second, integrate the spatial context into segmentation by fully leveraging feature interweaving among tasks.} The spatial context plays a significant role in improving the quality of segmentation. It is plentiful in the box regression sub-task but insufficient in others. To make other sub-tasks location-aware, we propose the information flows to deliver the spatial context from the box regression task to others and integrate it by feature interweaving. However, the absence of multi-stage features in thing and stuff segmentation makes it inconvenient to absorb the spatial context. To solve this dilemma, we design four parallel sub-networks for four sub-tasks in the framework, enabling the model to leverage feature interweaving among tasks.

The overall design fully leverages the spatial context, bridges all the tasks in panoptic segmentation by integrating features among them, and builds a global view for the image scene, leading to better refinement of features, more robust representations for image segmentation, and higher prediction results.


\textbf{Our contributions} are three-fold:
\begin{itemize}
\item In this paper, we present a new unified framework for panoptic segmentation. Our framework is built on the one-stage detector RetinaNet, which facilitates feature interweaving in pixel-level. 
\item Based on the proposed framework, we design four parallel sub-networks to refine sub-task features. Among the sub-networks, we propose the spatial information flows to bridge all sub-tasks by making them location-aware. Our framework is denoted as {\em SpatialFlow}. 
\item We perform a detailed ablation study on various components of SpatialFlow. Extensive experimental results show that SpatialFlow achieves state-of-the-art results, which are $\textbf{47.9}$ PQ and $\textbf{62.5}$ PQ on COCO~\cite{coco} and Cityscapes~\cite{cityscapes} panoptic benchmarks.
\end{itemize}

The rest of our paper is organized as follow: In Section~\ref{sec2}, we briefly revisit recent progress related to this paper; in Section~\ref{sec3}, we first present the proposed unified framework for panoptic segmentation based on RetinaNet, then we illustrate all the details of the designed parallel sub-networks and the spatial information flows; in Section~\ref{sec4}, \ref{sec5}, \ref{sec6}, we present all details and results of the experiments, analyze the effect of each component, and make further discussions; finally, we conclude the paper in Section~\ref{sec7}.

\section{Related Works} \label{sec2}
After the pioneering application of AlexNet~\cite{alexnet} on the ImageNet datasets~\cite{imagenet}, deep learning methods have come to dominate computer vision. These methods have dramatically improved the state-of-the-art in many vision tasks, including image recognition\cite{alexnet, yu2014high, vgg, googlenet, resnet, zhou2016fine, yu2019hierarchical}, image retrieval~\cite{yu2014learning, wang2015deep}, metric learning~\cite{hoffer2015deep, yu2016deep, oh2016deep}, object detection~\cite{rcnn, fastrcnn, fasterrcnn}, image segmentation~\cite{dilatedconv, unet, maskrcnn}, human pose estimation~\cite{cao2017realtime, hong2015multimodal}, and many other tasks. Our work builds on prior works in object detection and image segmentation. We apply multi-task learning~\cite{zhangoverview, kendall2018multi} in our model, which makes things and stuff segmentation tasks benefit each other and builds a global view for the image scene. Next, we review some works that are closest to our work as follows.

\subsection{Object Detection} 
Our community has witnessed remarkable progress in object detection. Works, such as~\cite{fastrcnn, fasterrcnn, rfcn}, tackled the detection problem by a two-stage approach. They first generated a number of object proposals as candidates, followed by a classification head and a regression head on each RoI. Numerous recent breakthroughs has been made, such as adjusting network structures~\cite{fpn, cascadercnn} and searching for better training strategies~\cite{snip, sniper, trident}. Another type of detector followed the single-stage pipeline, such as~\cite{yolo, ssd, retinanet}. They directly predict object categories and regress bounding box locations based on pre-defined anchors. Recently, researchers focus on improving the localization quality of one-stage detectors and propose anchor-free algorithms~\cite{garpn, fsaf, fcos, cornernet, centernet}. In ~\cite{retinanet}, the authors designed two parallel sub-networks for classification and regression, respectively. In this paper, SpatialFlow extends RetineNet by adopting the design of parallel sub-networks. 

\subsection{Instance Segmentation} 
Instance segmentation is a task that requires a pixel-level mask for each instance. Existing methods can be divided into two main categories, segmentation-based and region-based methods. Segmentation-based approaches, such as~\cite{instancelevelseg,categorylevelinstancelevelseg}, first generate a pixel-level segmentation map over the image and then perform grouping to identify the instance mask of each object. While region-based methods, such as~\cite{maskrcnn,panet,htc}, are closely related to object detection algorithms. They predict the instance masks within the bounding boxes generated by detectors. Region-based methods can achieve higher performance than their segmentation-based counterparts, which motivates us to resort to the region-based methods. In SpatialFlow, we adopt a thing branch upon RetinaNet for thing segmentation.
\begin{figure*}
  \centering
  \includegraphics[width=0.97\textwidth]{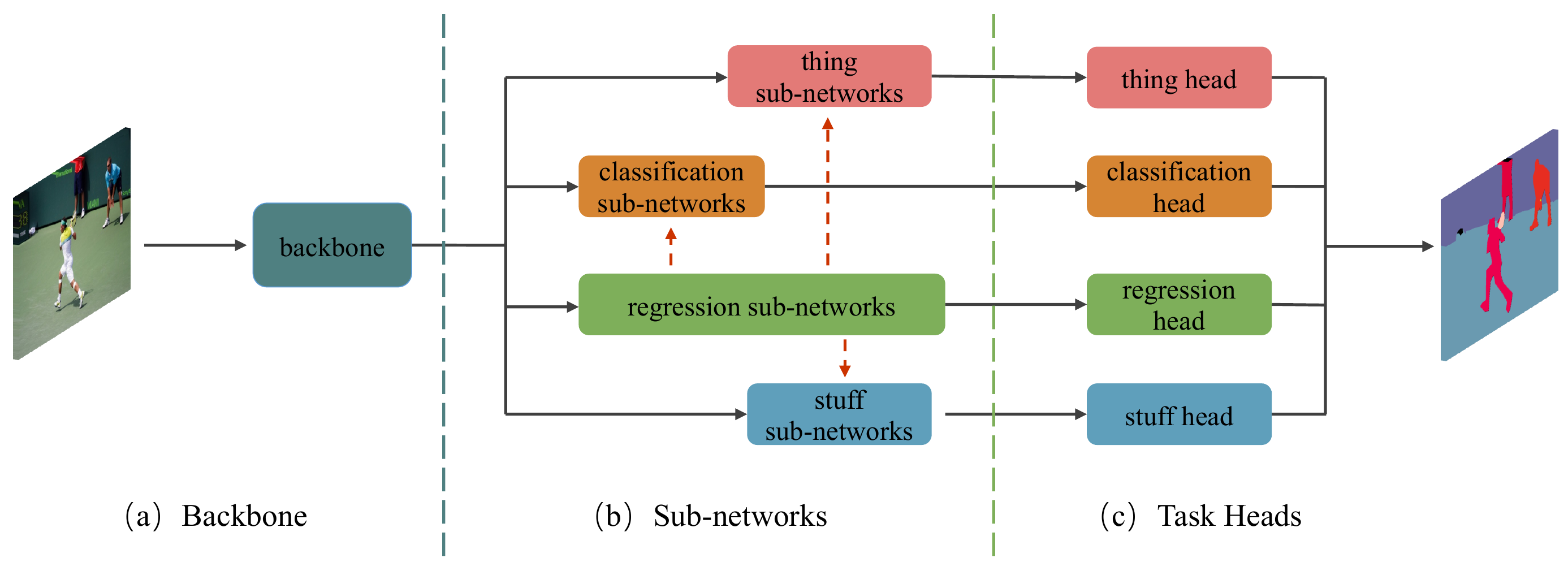}
  \caption{An illustration of the overall architecture. The SpatialFlow consists of three parts: (a) Backbone with FPN. (b) Four parallel sub-networks: We propose the spatial information flow and feature fusion among tasks in this part. The spatial flows are illustrated as orange dashed arrows, and the feature fusion is not shown in this figure for an elegant presentation; (c) Four heads for specific tasks: The classification head and regression head predict detection box together for thing head. The final result of SpatialFlow is a combination of the detected boxes and the outputs of thing head and stuff head.}
   \label{fig1}
\end{figure*}

\subsection{Semantic Segmentation} 
Fully convolutional networks are essential to semantic segmentation~\cite{fcn}, and its variants achieve state-of-the-art results on various segmentation benchmarks. It has been proven that contextual information plays a vital role in segmentation~\cite{contextrole}. A bunch of works followed this idea: dilated convolution~\cite{dilatedconv} was invented to keep feature resolution and maintain contextual details; Deeplab series~\cite{deeplab,deeplabv3} proposed Atrous Spatial Pyramid Pooling (ASPP) to capture global and multi-scale contextual information; PSPNet~\cite{pspnet} used spatial pyramid pooling to collect contextual priors; the encoder-decoder networks~\cite{unet,learningdeconv} are designed to capture contextual information in encoder and gradually recover the details in decoder. Our SpatialFlow, built upon FPN~\cite{fpn}, uses an encoder-decoder architecture for stuff segmentation to capture the contextual information. We take the spatial context of object detection into consideration and build a connection for thing and stuff segmentation.

\subsection{Panoptic Segmentation} 
The panoptic segmentation task was proposed in~\cite{panopticsegmentation}, where the authors provided a baseline method with two separate networks, then used a heuristic post-processing method to merge two outputs. Later, Li et al.~\cite{weaklypanoptic} followed this task and introduced a weakly- and semi-supervised panoptic segmentation method. Recently, several unified frameworks have been proposed. De Geus et al.~\cite{jsisnet} used a shared backbone for both things and stuff segmentation, while Li et al.~\cite{tascnet} took a step further by considering things and stuff consistency and proposed a unified network named TASCNet. Kirillov et al.~\cite{panopticfpn} introduced PanopticFPN by endowing Mask R-CNN~\cite{maskrcnn} with a stuff branch, which ignores the connection between things and stuff. Li et al.~\cite{aunet} aimed to capture the connection by utilizing the attention module. To solve the conflicts in the result merging process, Liu et al.~\cite{panopticranking} designed a spatial ranking module. Also, Xiong et al.~\cite{upsnet} proposed a parameter-free panoptic head to resolve the conflicts. Thinking differently, Yang et al. presented a single-shot approach for panoptic segmentation. However, most of these methods ignored to highlight the significance of the spatial features. Our SpatialFlow proposes the information flows to enable all tasks to be location-aware, which helps build a panoptic view for image segmentation.

\section{SpatialFlow} \label{sec3}
Object location is one of the key factors when building a global view for panoptic segmentation. While recent works~\cite{jsisnet,tascnet,panopticfpn,upsnet,panopticranking} for panoptic segmentation focus on how to combine the outputs of segmentation tasks properly but ignore to highlight the significance of the object location in the training phase. In this work, we propose a new unified framework, SpatialFlow, which enables all sub-tasks to be location-aware. Our SpatialFlow is conceptually simple: RetinaNet~\cite{retinanet} with two added sub-networks and two extra heads for thing and stuff segmentation. More importantly, we add multi-stage spatial information flows among the sub-networks.

We begin by reviewing the RetinaNet detector. RetinaNet is one of the most successful fully convolutional one-stage detectors. It consists of three parts: backbone with FPN~\cite{fpn}, two parallel sub-networks, and two task-specific heads for box classification and regression. In SpatialFlow, we adopt the main network structure of RetinaNet. We illustrate the sketch of our framework in Figure~\ref{fig1}. 
\begin{figure*}
  \centering
  \includegraphics[width=1.\textwidth]{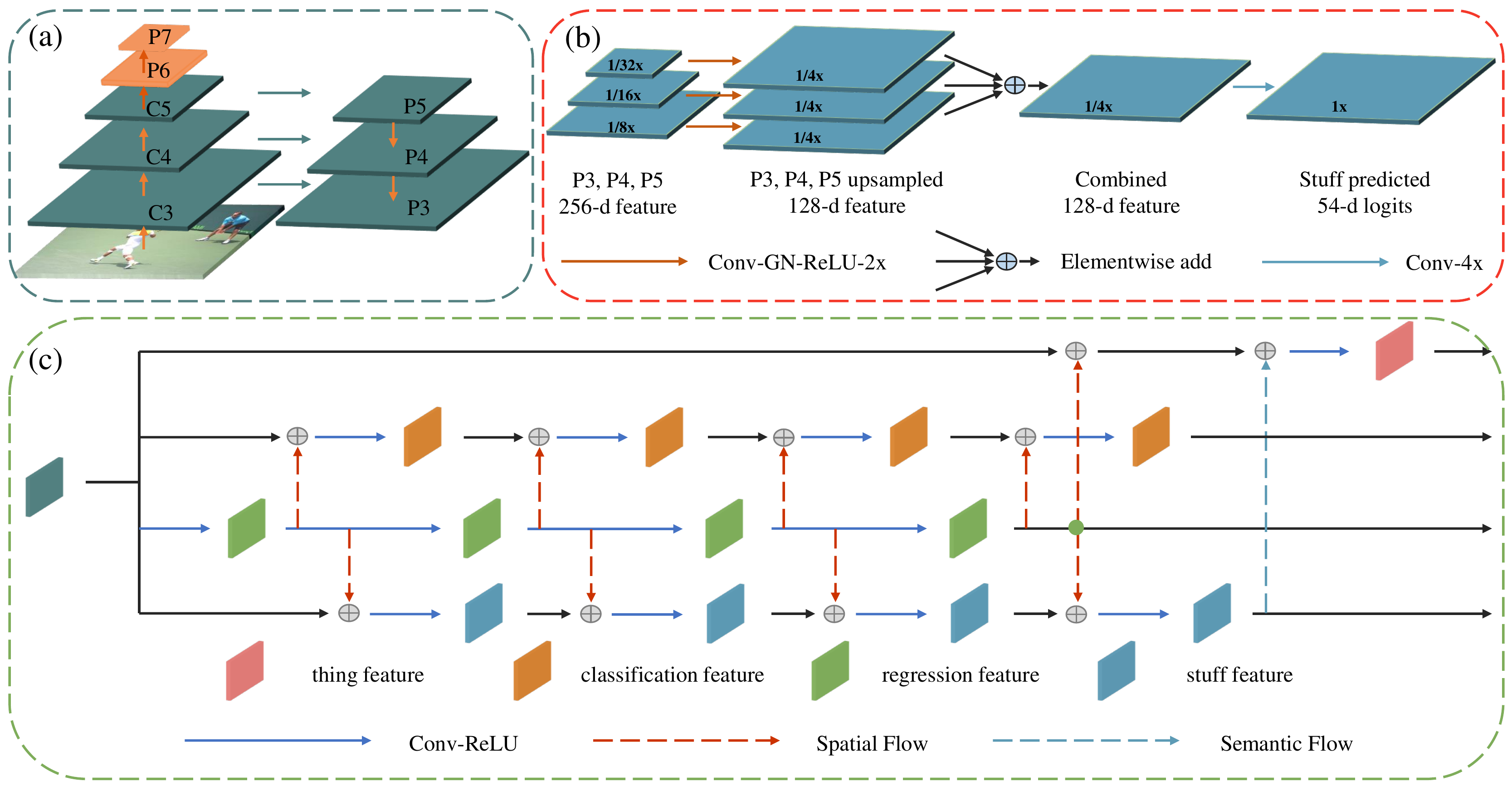}
  \caption{The designs for each part in SpatialFlow. In the dashed rectangle (a), we show the output features of FPN, which are the features named $\{P_3, P_4, P_5, P_6, P_7\}$. In the dashed rectangle (b), we present the architecture of the stuff head. More importantly, all the information flows in sub-networks are illustrated in the dashed box (c).}
   \label{fig2}
\end{figure*}

\subsection{Naive Implementation} \label{sec3.1}
As we discussed in Section~\ref{intro}, RetinaNet shows its merits in pixel-level feature integration, which is beneficial for segmentation tasks. To develop a unified framework for panoptic segmentation based on RetinaNet, the most naive way is to add one thing head and one stuff head upon FPN features to enable thing and stuff segmentation. In this section, we introduce the naive implementation of the unified framework, which ignores the task feature refinement and the integration of box locations like previous methods~\cite{panopticfpn,panopticranking,upsnet} but built on RetinaNet. Next, we will introduce the detailed design of each element in the naive implementation.

\subsubsection{Backbone} 
We adopt the same backbone structure as RetinaNet. The backbone contains FPN, whose outputs are five levels of features named $\{P_3, P_4, P_5, P_6, P_7\}$ with a downsample rate of $8$, $16$, $32$, $64$, $128$ respectively. In FPN, all features have $256$ channels. We show the details in Figure~\ref{fig2} (a). Following~\cite{retinamask}, we treat these features differently against various tasks: we use all the five levels to predict the bounding boxes in detection but only send $\{P_3, P_4, P_5\}$ to thing and stuff segmentation.
\\[0.5mm]
\subsubsection{RetinaNet-based sub-networks} 
We present the parallel sub-networks in RetinaNet - classification sub-network (cls sub-net for short) and regression sub-network (reg sub-net for short). The operations in these sub-networks, which transform the output features of FPN to the inputs of downstream heads, can be formulated as follows:
\begin{equation} \label{eq1}
P_{reg_{i, j}} = \phi(P_{reg_{i, j-1}}), \quad P_{cls_{i, j}} = \phi(P_{cls_{i, j-1}}).
\end{equation}
Here, $i$ represents the level index of FPN levels, $j$ is the layer stage index in sub-networks, and $\phi$ denotes to a network block that contains a $3\times3$ convolution layer and a ReLU layer. In the cls and reg sub-networks, $i \in \{3, 4, 5, 6, 7\}$, $j \in \{1, 2, 3, 4\}$, and we have $P_{cls_{i, 0}} =  P_{reg_{i, 0}} = P_i$, while $i \in \{3, 4, 5\}$ for thing and stuff segmentation.
\\[0.5mm]
\subsubsection{Task-specific heads} 
As illustrated in Figure~\ref{fig1} (c), we apply four heads for box classification, box regression, thing segmentation, and stuff segmentation, respectively. In the classification and the regression head, the final outputs of the detection can be obtained by $O_{cls_{i}} = \psi(P_{cls_{i, 4}}), O_{reg_{i}} = \phi(P_{reg_{i, 4}})$, where $O_{cls_{i}}$ and $O_{reg_{i}}$ represent the outputs of the classification head and the regression head in the FPN level $i$. We implement one $3 \times 3$ convolution layer $\phi$ on the outputs of classification and regression sub-nets. For the thing head, we apply it to each predicted box and adopt the same design as Mask R-CNN~\cite{maskrcnn}. For each RoI feature, $O_{RoI_{k}} = \psi(\zeta(\phi(P_{RoI_{k}}))$, where $O_{RoI_{k}}$ is the output of the $k$-th predicted box, $\phi$ represents for four $3 \times 3$ convolution layers with ReLU, one $2 \times 2$ stride 2 deconvolution layer with ReLU is denoted as $\zeta$, and $\psi$ is a $1 \times 1$ output convolution layer. After the stuff sub-net, we obtain three levels of feature maps with scales of $1/8$, $1/16$, $1/32$ of the original image. We perform upsampling on each feature map gradually by blocks, each of which contains a $3 \times 3$ convolution layer, a group norm~\cite{groupnorm} layer, a ReLU layer, and a $2\times$ bilinear upsampling operation. All the features are upsampled to the scale of $1/4$, which are then element-wise summed. A final $1 \times 1$ convolution layer, a $4\times$ bilinear upsampling operation, and a softmax are applied to get the segmentation result. The stuff head is shown in Figure~\ref{fig2} (b) with details. To generate the final output of SpatialFlow, we first perform a heuristic post-processing method~\cite{panopticfpn} to merge the results of thing and stuff segmentation, then fill the unassigned area in the merged map with the predicted boxes' locations and categories.

We show the key components of the proposed unified framework. The adaptation of RetinaNet~\cite{retinanet} enables the feature in pixel-level before performing segmentation tasks. There remain obstacles preventing the unified framework from building a global view for the image scene, e.g., lack of feature intersection between things and stuff. The naive implementation also has practicality problems regarding the refinement of the FPN feature for things and stuff segmentation. To further improve the quality of learned features for things and stuff segmentation and strengthen the intersection between things and stuff in the image, we propose two techniques: adding things and stuff parallel sub-networks and proposing spatial information flows.

\subsection{Thing and stuff parallel sub-networks}
In RetinaNet, the parallel sub-networks refine the FPN features with multi-stage convolution layers, which transform the FPN features to task-specific features and lead to better performance. But there is no refinement for the input features in thing and stuff segmentation in the naive implementation. In this section, we apply the same mechanism to these two segmentation tasks. Moreover, the created multi-stage features facilitate the delivery of the spatial context from box regression task to others. We show the details of this part in Figure~\ref{fig2} (c).

In this section, we propose to add two additional sub-networks - thing sub-network and stuff sub-network. We adopt the similar structure as in cls and reg sub-networks. Until now, there are four parallel sub-networks between the FPN and the task-specific heads. We present the modifications in thing and stuff sub-networks below:
\begin{equation} \label{eq2}
\begin{split}
&P_{thing_{i, j}} = \phi(P_{thing_{i, j-1}}); \\
&P_{stuff_{i, j}} = \phi(P_{stuff_{i, j-1}}).
\end{split}
\end{equation}
Where $P_{thing_{i, 0}} =  P_{stuff_{i, 0}} = P_i$. As the dominated features in thing and stuff segmentation are different, the number of stages required by the sub-networks depends on tasks. More stages are needed in stuff segmentation than in thing segmentation to do feature refinement. We conjecture that the reason for the phenomenon is that pixel-level features are more sensitive to the details than instance-level features. In the final version, we adopt four stages in stuff sub-network and keep only one in thing sub-network. This setting gives the best performance of segmentation. In each stage, we implement a $3\times3$ convolutional layer and a ReLU layer. We illustrate the overall structure of sub-networks in Figure~\ref{fig2} (c). And the experimental results for the number of stages in thing and stuff sub-networks can be found in Table~\ref{tab7a} and Table~\ref{tab7b}.

\subsection{Spatial information flows}
As illustrated in Figure~\ref{fig0}, all sub-tasks in our proposed panoptic segmentation framework are related to the locations of objects. The box location information is implied in the multi-stage feature representations of the box regression sub-network. We propose the spatial information flows to support feature refinement in sub-networks. The spatial information flows can make other sub-tasks aware of box locations. Furthermore, adding the semantic feature in the thing segmentation has been proved to be effective in HTC~\cite{htc}. We also add a semantic flow that adopts a $3 \times 3$ convolution layer to transform the stuff feature to the thing feature. It brings slight improvements in our SpatialFlow, as shown in Table~\ref{tab8a} and Table~\ref{tab8b}. Then we display the detailed structure of the spatial flows in Figure~\ref{fig2} (c). They can be implemented as follows:
\begin{equation} \label{eq3}
\begin{split}
&P_{reg_{i, j}} =  \phi(P_{reg_{i, j-1}}); \\
&P_{cls_{i, j}} =  \phi(P_{cls_{i, j-1}} + \psi(P_{reg_{i, j}})); \\
&P_{stuff_{i, j}} =  \phi(P_{stuff_{i, j-1}} + \psi(P_{reg_{i, j}})); \\
&P_{thing_{i, 1}} = \phi(P_i + \zeta(P_{stuff_{i, 4}}), \psi_{offset}(P_{reg_{i, 4}} + P_i)); \\
&P_{reg_{i, 0}} = P_{cls_{i, 0}} = P_{thing_{i, 0}} = P_{stuff_{i, 0}} = P_i.
\end{split}
\end{equation}
Here, $P_{reg_{i, 0}} = P_{cls_{i, 0}} = P_{thing_{i, 0}} = P_{stuff_{i, 0}} = P_i$ and $\psi$ denotes an adaptation convolution from box regression task to others; $\zeta$ denotes an adaptation convolution from stuff sub-net to thing sub-net. We use a $3 \times 3$ convolution layer for both $\psi$ and $\zeta$. All features have 256 channels in this part.

Moreover, to make a fair comparison with UPSNet~\cite{upsnet} on COCO, we introduce deformable convolutions~\cite{deformable} layers to the sub-networks. We further adopt a method to incorporate the spatial context into deformable convolution more appropriately. We first combine the spatial information flow and the task-specific feature, then use the combined feature to generate the offsets for the deformable convolution on the task-specific sub-networks. The process can be formulated as follow:
\begin{equation} \label{eq6}
\begin{split}
&P_{reg_{i, j}} = \phi(P_{reg_{i, j-1}}); \\
&P_{cls_{i, j}} = \phi_{dcn}(P_{cls_{i, j-1}}, \psi_{offset}(P_{reg_{i, j}} +P_{cls_{i, j-1}})); \\
&P_{stuff_{i, j}} = \phi_{dcn}(P_{stuff_{i, j-1}}, \psi_{offset}(P_{reg_{i, j}} + P_{stuff_{i, j-1}})); \\
&P_{thing_{i, 1}} = \phi(P_i + \zeta(P_{stuff_{i, 4}}), \psi_{offset}(P_{reg_{i, 4}} + P_i)); \\
&P_{reg_{i, 0}} = P_{cls_{i, 0}} = P_{thing_{i, 0}} = P_{stuff_{i, 0}} = P_i.
\end{split}
\end{equation}
In the equation, $\phi_{dcn}$ represents a deformable convolution layer, $\psi_{offset}$ means an adaptation convolution layer, which generates offsets for the deformable convolution. {\em Unless specified, we do not adopt the setting with deformable convolution}.

\section{Experiments} \label{sec4}
\subsection{Dataset and Evaluation metric}
\subsubsection{Dataset} 
We evaluate our model on both COCO~\cite{coco} and Cityscapes~\cite{cityscapes}. COCO consists of 80 things and 53 stuff classes. We use the 2017 data splits with 118k/5k/20k {\em train}/{\em val}/{\em test} images. We use {\em train} split for training, and report leision and sensitive studies by evaluating on {\em val} split. For our main results, we report our panoptic performance on the {\em test-dev} split. Cityscapes has $5k$ high-resolution images with fine pixel-accurate annotations: $2975$ train, $500$ val, and $1525$ test. There are $19$ classes on Cityscapes, $8$ with instance-level masks. For all experiments on Cityscapes, we report our performance on {\em val} split with $11$ stuff classes and $8$ things classes.
\\[0.5mm]
\subsubsection{Evaluation metric} 
We adopt the {\em panoptic quality} (PQ) as the metric. As proposed in~\cite{panopticsegmentation}, PQ can be formulated as follow:
\begin{equation} \label{eq5}
PQ = \underbrace{\frac {\sum_{(p,g) \in TP} IoU(p, g)}{|TP|}}_{\text{segmentation quality (SQ)}} \times \underbrace{\frac {|TP|}{|TP| + \frac 12 |FP| + \frac 12 |FN|}}_{\text{recognition quality (RQ)}}
\end{equation}
where $p$ and $g$ are predicted and ground truth segments, TP (true positives), FP (false positives), and FN (false negatives) represent matched pairs of segments ($IoU(p, g) > 0.5$), unmatched predicted segments, and unmatched ground truth segments, respectively. Besides, PQ can be explained as the multiplication of a segmentation quality (SQ) and a recognition quality (RQ). We also use SQ and RQ to measure the performance in our experiments.

\subsection{Implementation Details}
\subsubsection{Training} 
As a unified framework for panoptic segmentation, there are four different losses for SpatialFlow to optimize during the training stage. The loss function can be formulated as follow:
\begin{equation} \label{eq4}
\mathcal{L} = (\mathcal{L}_{cls} + \mathcal{L}_{reg} + \mathcal{L}_{thing}) + \lambda \cdot \mathcal{L}_{stuff}
\end{equation}
where $\mathcal{L}_{cls}$, $\mathcal{L}_{reg}$ and $\mathcal{L}_{thing}$ belong to the thing segmentation task, and $\mathcal{L}_{stuff}$ is the loss of the stuff segmentation. We add a hyper-parameter $\lambda$ to balance the losses between thing and stuff segmentation. We implement our SpatialFlow with a toolbox~\cite{mmdetection} based on PyTorch~\cite{pytorch}. 

We inherit all the hyper-parameters from RetinaNet except that we set the threshold of NMS to $0.4$ when generating proposals during training. For thing prediction, we add the ground truth boxes to the proposals set and run the thing head for all proposals. For training strategies, we fix the batch norm layer in the backbone and train all models over $4$ GPUs with a total of 8 images per minibatch. On MS-COCO~\cite{coco}, we use the training strategy of training longer that adopted by RetinaNet(1.5$\times$)~\cite{retinanet} and RetinaMask(2$\times$)~\cite{retinamask}. All models are trained for $20$ epochs with an initial learning rate of $5 \times 10^{-3}$, which is decreased by $10$ after $16$ and $19$ epochs; on Cityscapes~\cite{cityscapes}, we set the initial learning rate as $1.25 \times 10^{-2}$ and borrow the number of iterations from~\cite{panopticfpn}. Unless specified, we resize the shorter edge of the image to $800$ pixels on COCO, while on Cityscapes, we adopt $512 \times 1024$ image crops after scaling each image by $0.5$ to $2.0\times$. As Kirillov et al.~\cite{panopticfpn} did, we also predict a particular `other' class for all things categories in stuff head on COCO benchmark.
\\[0.5mm]
\subsubsection{Inference} 
Our model follows a pipeline in the inference stage: (1) generate the detection results; (2) obtain the maps of thing and stuff segmentation; (3) merge the two outputs to form a panoptic segmentation map; (4) fill the unassigned area in the result with the detected boxes and its categories. In detection, we set the threshold of NMS to $0.4$ for each class separately, and choose the top-$100$ scoring bounding boxes to send to thing head. During merging, we first ignore the stuff regions labeled `other'; then we resolve the overlap problem between instances based on their scores, and merge the thing and stuff map in favor of things; at last, we fill the unassigned area with detection boxes in the result segmentation map to form the final output. For the hyper-parameters of SpatialFlow on in the inference stage, we fixed the confidence score threshold for the instance masks as $0.37$, set the overlap threshold of instance masks as $0.37$, and set the area limit threshold of stuff regions as $4900$. When performing integration with detection box results, we propose a new hyper-parameter, which is the overlap between the detection box and the unassigned area in the segmentation map. We fix the threshold of the box overlap as $0.6$.  For the hyper-parameters on Cityscapes, we modify the overlap threshold of the instance masks to $0.25$ and change the area limit threshold of stuff regions to $2048$.
\begin{table*}
  \centering
   \caption{Comparison with the state-of-the-art methods on COCO 2017 {\em test-dev} split. We only compare with the state-of-the-art methods that without deformable convolutions here. }
  \begin{tabular}{cccccccccc}
    \toprule
    model & backbone & PQ & PQ$^{Th}$ & PQ$^{St}$ & SQ & RQ \\
    \midrule
    \midrule
    JSIS-Net~\cite{jsisnet} & ResNet-50 & 27.2 & 29.6 & 23.4 & 71.9 & 35.9 \\
    DeeperLab~\cite{deeperlab} & Xception-71 & 34.3 & 37.5 & 29.6 & 77.1 & 43.1 \\
    PanopticFPN~\cite{panopticfpn} & ResNet-101-FPN & 40.9 & 48.3 & 29.7 & - & -  \\
    OANet~\cite{panopticranking} & ResNet-101-FPN & 41.3 & \textbf{50.4} & 27.7 & - & - \\
    SSAP~\cite{ssap} & ResNet-101-FPN & 36.9 & 40.1 & 32.0 & - & - \\
    \textbf{SpatialFlow} & ResNet-101-FPN & \textbf{42.9} & 49.5 & \textbf{33.0} & \textbf{78.8} & \textbf{52.3} \\
    \bottomrule
  \end{tabular}
   \label{tab2}
\end{table*}
 \begin{table*}
  \centering
    \caption{Comparison with the state-of-the-art methods on COCO 2017 {\em test-dev} split. In this table, we report our results with deformable convolution and multi-scale strategy. The top 3 rows contain results of top 3 models taken from the official leaderboard of COCO 2018 Panoptic Segmentation Challenge.}
  \begin{tabular}{ccccccccc}
    \toprule
    model & backbone & PQ & PQ$^{Th}$ & PQ$^{St}$ & SQ & RQ \\
    \midrule
    \midrule
    Megvii (Face++) & ensemble model & 53.2 & 62.2 & 39.5 & 83.2 & 62.9 \\
    Caribbean & ensemble model & 46.8 & 54.3 & 35.5 & 80.5 & 57.1\\
    PKU 360 & ResNeXt-152-DCN & 46.3 & 58.6 & 27.6 & 79.6 & 56.1 \\
    \midrule
    AdaptIS~\cite{adaptis} & ResNeXt-101 & 42.8 & 50.1 & 31.8 & - & - \\
    AUNet~\cite{aunet} & ResNeXt-152-DCN &  46.5  & \textbf{55.9} & 32.5 & 81.0 & 56.1\\
    UPSNet~\cite{upsnet} & ResNet-101-DCN &  46.6 & 53.2 & 36.7 & 80.5 & 56.9 \\
    SOGNet~\cite{sognet} & ResNet-101-DCN & 47.8 & - & - & 80.7 & 57.6 \\
    \textbf{SpatialFlow} & ResNet-101-DCN &  \textbf{47.9} & 54.5 & \textbf{38.0} & \textbf{81.7} & \textbf{57.6}\\
    \bottomrule
  \end{tabular}
  \label{tab3}
\end{table*}
\begin{table}
  \centering
  \caption{Comparison with the state-of-the-art methods on Cityscapes {\em val} split. In this table, `-R101' represents that the backbone is ResNet-101 and `-X101` for ResNeXt-101~\cite{resnext}; `-COCO' means using COCO pretrained model; `-M' is the multi-scale testing. }
  \begin{tabular}{cccc}
    \toprule
    model & PQ & PQ$^{Th}$ & PQ$^{St}$ \\
    \midrule
    \midrule
    PanopticFPN-R101~\cite{panopticfpn} & 58.1 & 52.0 & 62.5 \\
    AUNet-R101~\cite{aunet} & 59.0 & 54.8 & 62.1 \\
    TASCNet-R101-COCO~\cite{tascnet} & 59.2 & 56.0 & 61.5 \\
    UPSNet-R101-COCO-M~\cite{upsnet} & 61.8 & 57.6 & 64.8 \\
    SSAP-R101-M~\cite{ssap} & 61.1 & 55.0 & - \\
    AdaptIS-X101-M~\cite{adaptis} & 62.0 & \textbf{58.7} & 64.4 \\
    \textbf{SpatialFlow-R101} & 59.6 & 55.0 & 63.1 \\
    \textbf{SpatialFlow-R101-COCO-M} & \textbf{62.5} & 56.6 & \textbf{66.8} \\
    \bottomrule
  \end{tabular}
  \label{tab4}
\end{table}
\subsection{Main Results}
In this section, we compare our SpatialFlow with the state-of-the-art methods in panoptic segmentation. We show all the main results in Table~\ref{tab2}, Table~\ref{tab3}, and Table~\ref{tab4}. SpatialFlow achieves state-of-the-art results on both COCO~\cite{coco} and Cityscapes~\cite{cityscapes} panoptic benchmark.

\subsubsection{MS COCO} 
To make a fair comparison, we report the results in Table~\ref{tab2} and Table~\ref{tab3}, where the experiment settings are different. In Table~\ref{tab2}, we present the prediction results without bells and whistles. With a single ResNet-101-FPN backbone, our SpatialFlow can achieve $\textbf{42.9}$ PQ on COCO {\em test-dev} split, which outperforms PanopticFPN~\cite{panopticfpn} by $2.0$ PQ and OANet~\cite{panopticranking} by $1.6$ PQ. More importantly, SpatialFlow achieves a new state-of-the-art performance on PQ$^{St}$, $\textbf{33.0}$ PQ, which outperforms other models by a large margin ($3.3$ PQ and $5.3$ PQ respectively). The results demonstrate the effectiveness of integrating the spatial features in pixel-level, which significantly impact stuff segmentation. However, SpatialFlow is lagging behind OANet~\cite{panopticranking} in PQ$^{Th}$. In OANet, the authors focus on solving the overlapping problem of instances when rendering the thing results to the final panoptic result. SpatialFlow applies a simple method to this problem, causing the inferior performance in PQ$^{th}$.

Then we apply deformable convolution~\cite{deformable} to both backbone and sub-networks and report its results with the multi-scale strategy in Table~\ref{tab3}. When training, the scales of short edges are randomly sampled from $[400, 1400]$, and the scales of long edges are fixed as 1600. For inference, we feed multi-scale images to SpatialFlow, and the scales are $(1500, 1000)$, $(1800, 1200)$, and $(2100, 1400)$ with horizontal flip. We achieve $\textbf{47.9}$ PQ, which is the state-of-the-art result on COCO panoptic benchmark. As shown in Table~\ref{tab3}, our method outperforms the runner-up of the COCO 2018 challenge by $1.1$ PQ with a single model, demonstrating the effectiveness of SpatialFlow. Although AUNet outperforms our method in PQ$^{th}$, they use ResNeXt-152-DCN as their backbone. In fact, with a stronger backbone (ResNeXt-101-DCN) and model ensemble, our method can achieve $50.2$ PQ on COCO {\em test-dev} split. 

\subsubsection{Cityscapes} 
We also report the results under different experiment settings in Table~\ref{tab4}. Without the COCO pre-trained model, SpatialFlow can achieve $\textbf{59.6}$ PQ on Cityscapes {\em val} split, which is $1.5$ PQ and $0.6$ PQ higher than PanopticFPN~\cite{panopticfpn} and AUNet~\cite{aunet} respectively. With the COCO pre-trained model, SpatialFlow can achieve $\textbf{62.5}$ PQ on Cityscapes with multi-scale testing, which is $0.7$ PQ higher than UPSNet~\cite{upsnet} under the same setting. SpatialFlow outperforms all other methods in terms of PQ$^{st}$ while getting inferior performance on PQ$^{th}$ comparing to UPSNet and AdaptIS. We conjecture that the phenomenon is caused by the inferior detection performance of RetinaNet on Cityscapes. To obtain the result of $62.5$ PQ on Cityscapes {\em val} split, we first replace the convolution layers in stuff with deformable convolutions as UPSNet does, then we follow the steps below: (1) Finetune the COCO pre-trained model. As the number of things and stuff classes in Cityscapes is smaller than the number in COCO, $11/8$ vs. $80/53$, we have to finetune the layers that related to the number of classes. We freeze the rest layers and use a learning rate of $2.5 \times 10^{-3}$ to train for $2$ epochs. (2) Train the finetuned model as the standard SpatialFlow does. (3) Apply the multi-scale testing trick. The scales that we use in Cityscapes are $(2304, 1152)$, $(2432, 1216)$, $(2560, 1280)$, and $(2688, 1344)$ with horizontal flip.
\begin{table}
  \centering
  \caption{\textbf{Loss Balance}: The results of the baseline model on COCO {\em val} for different values of $\lambda$ based on ResNet-50 with image size of 600px. The proper $\lambda$ brings a large gain.}
  \begin{tabular}{lcccccc}
    \toprule
    $\lambda$ & 1.0 & 0.75 & 0.5 & 0.3 & \textbf{0.25} & 0.2\\
    \midrule
    \midrule
    PQ & 37.5 & 38.2 & 38.8 & 39.1 & \textbf{39.3} & 39.0 \\
    PQ$^{Th}$ & 41.8 & 43.0 & 44.0 & 44.5 & \textbf{45.1} & 44.9\\
    PQ$^{St}$ & 30.9 & 31.1 & 31.0 & 30.9 & \textbf{30.5} & 30.0\\
    \bottomrule
  \end{tabular}
  \label{tab6}
\end{table}
\begin{table*}[!t]
\begin{minipage}[!t]{\columnwidth}
 \caption{\textbf{Contribution of Components}: Ablation results on COCO {\em val} split with ResNet-50. Both sub-networks and spatial flows bring significant gains based on the baseline model.}
\label{tab5a}
  \centering
  \setlength{\tabcolsep}{2.0mm}{
  \begin{tabular}{ccccc}
    \toprule
    sub-nets & spatial-flows & PQ & PQ$^{Th}$ & PQ$^{St}$ \\
    \midrule
    \midrule
     &  & 39.7 & 46.0 & 30.2 \\
    \checkmark &  & 40.3 & 46.2 & 31.4 \\
     & \checkmark & 40.2 & 46.5 & 30.7 \\
     \checkmark & \checkmark & \textbf{40.9} & \textbf{46.8} & \textbf{31.9} \\
    \bottomrule
  \end{tabular}}
  \end{minipage}
\begin{minipage}[!t]{\columnwidth}
  \caption{\textbf{Contribution of Components}: Ablation results on Cityscapes {\em val} split with ResNet-50. Similar gains can be obtained on Cityscapes.}
\label{tab5b}
  \centering
  \setlength{\tabcolsep}{2.0mm}{
\begin{tabular}{ccccc}
    \toprule
    sub-nets & spatial-flows & PQ & PQ$^{Th}$ & PQ$^{St}$ \\
    \midrule
    \midrule
     &  & 57.3 & 53.5 & 60.0 \\
    \checkmark &  & 57.5 & 53.6 & 60.3 \\
     & \checkmark & 58.0 & 54.3 & 60.8 \\
     \checkmark & \checkmark & \textbf{58.6} & \textbf{54.9} & \textbf{61.4} \\
    \bottomrule
  \end{tabular}}
  \end{minipage}
\end{table*}
\begin{table*}[!t]
\begin{minipage}[!t]{\columnwidth}
\caption{\textbf{Design of Sub-networks}: Ablation results on number of stages in thing sub-network. Only one stage is needed to refine the FPN feature for the input of the thing head.}
\label{tab7a}
  \centering
  \setlength{\tabcolsep}{3mm}{
\begin{tabular}{cccc}
    \toprule
    num stages & PQ & PQ$^{Th}$ & PQ$^{St}$ \\
    \midrule
    \midrule
    0 & 39.7 & 46.0 & 30.2 \\
     \textbf{1} & \textbf{39.9} & \textbf{46.3} & \textbf{30.3} \\
     2 & 39.7 & 46.0 & 30.2 \\
     3 & 39.9 & 46.2 & 30.3 \\
     4 & 39.7 & 45.9 & 30.3 \\
    \bottomrule
  \end{tabular}}
  \end{minipage}
\begin{minipage}[!t]{\columnwidth}
\caption{\textbf{Design of Sub-networks}: Results on number of stages in stuff sub-network. More stages bring more gains. Input feature of the stuff head need to be fully refined by sub-network.}
\label{tab7b}
  \centering
  \setlength{\tabcolsep}{3mm}{
\begin{tabular}{cccc}
    \toprule
    num stages & PQ & PQ$^{Th}$ & PQ$^{St}$ \\
    \midrule
    \midrule
    0 & 39.7 & 46.0 & 30.2 \\
     1 & 39.9 & 45.9 & 30.9 \\
     2 & 40.1 & 46.0 & 31.1 \\
     3 & 40.2 & 46.1 & 31.4 \\
     \textbf{4} & \textbf{40.3} & \textbf{46.2} & \textbf{31.5} \\
    \bottomrule
  \end{tabular}}
  \end{minipage}
\end{table*}
\begin{table*}[!t]
\begin{minipage}[!t]{\columnwidth}
\caption{\textbf{Spatial Flows}: The results of the spatial flows on COCO. Each row adds an extra component to the above.}
\label{tab8a}
  \centering
  \setlength{\tabcolsep}{4mm}{
\begin{tabular}{cccc}
    \toprule
    flows & PQ & PQ$^{Th}$ & PQ$^{St}$ \\
    \midrule
    - & 40.3 & 46.2 & 31.4 \\
     + reg-cls & 40.5 & 46.6 & 31.4 \\
     +reg-stuff & 40.7 & 46.3 & 32.0 \\
     +reg-thing & 40.7 & 46.4 & 31.8 \\
     +stuff-thing & \textbf{40.9} & \textbf{46.8} & \textbf{31.9} \\
    \bottomrule
  \end{tabular}}
  \end{minipage}
\begin{minipage}[!t]{\columnwidth}
\caption{\textbf{Spatial Flows}: The results of the spatial flows on Cityscapes. Each row adds an extra component to the above.}
\label{tab8b}
  \centering
  \setlength{\tabcolsep}{4mm}{
\begin{tabular}{cccc}
    \toprule
    flows & PQ & PQ$^{Th}$ & PQ$^{St}$ \\
    \midrule
    - & 57.5 & 53.6 & 60.3 \\
     + reg-cls & 58.0 & 55.1 & 60.1 \\
     +reg-stuff & 58.3 & 54.6 & 60.9 \\
     +reg-thing & 58.5 & 54.7 & 61.3 \\
     +stuff-thing & \textbf{58.6} & \textbf{54.9} & \textbf{61.4} \\
    \bottomrule
  \end{tabular}}
  \end{minipage}
\end{table*}
\subsection{Ablation Experiments}
We run a number of ablations to analyze the SpatialFlow. Unless specified, we use the naive implementation of SpatialFlow presented in Section~\ref{sec3.1} as our baseline model for all experiments in this section. We discuss the details below.
\\[3mm]
\textbf{Loss Balance.} We first investigate the best value of the hyper-parameter $\lambda$. We adopt the baseline model in this section. Table~\ref{tab6} shows the model results of using various $\lambda$ on COCO. We demonstrate the power of $\lambda$ and discover that the best value to balance the losses on COCO is $0.25$, with which the baseline model achieves 39.3 PQ with an image size of $600$px and earns a 1.8 PQ gain compared with $\lambda=1.0$. While for Cityscapes, we set $\lambda=1.0$ by following~\cite{panopticfpn}.
\\[3mm]
\textbf{Contribution of Components.} In this section, we evaluate the sub-networks and the spatial flows of SpatialFlow on both COCO and Cityscapes. The results are shown in Table~\ref{tab5a} and Table~\ref{tab5b} respectively. From the experiment results, we can see that both the sub-networks and the spatial flows demonstrate their contribution. The sub-networks improve PQ by $0.6$ points and $0.2$ points on COCO and Cityscapes. In particular, we obtain a significant gain on stuff ($1.2$ PQ on COCO) with the sub-networks as in which we refine the pixel-level feature before sending it to stuff head. For the spatial flows, they can improve the performance of things and stuff simultaneously by improving $0.5$ PQ and $0.7$ PQ on COCO and Cityscapes. Moreover, the spatial flows can bring further gains with the sub-networks compared with the obtained benefits on the baseline model. The results indicate that the integration of the spatial context can benefit from the feature refinement in sub-networks.
\\[3mm]
\textbf{Design of Sub-networks.} We search the best number of stages for thing and stuff sub-networks. We conduct experiments on the COCO dataset with ResNet-50 based on the baseline model. The results are shown in Table~\ref{tab7a} and Table~\ref{tab7b}. According to the results, we choose to add only one stage in thing sub-network and add four stages in stuff sub-network. We obtain 0.2 PQ and 0.6 PQ improvements with thing sub-network and stuff sub-network. The different number of blocks in sub-networks are related to the difference of the dominated feature in thing and stuff segmentation.
\begin{figure*}
  \centering
  \includegraphics[width=0.83\textwidth]{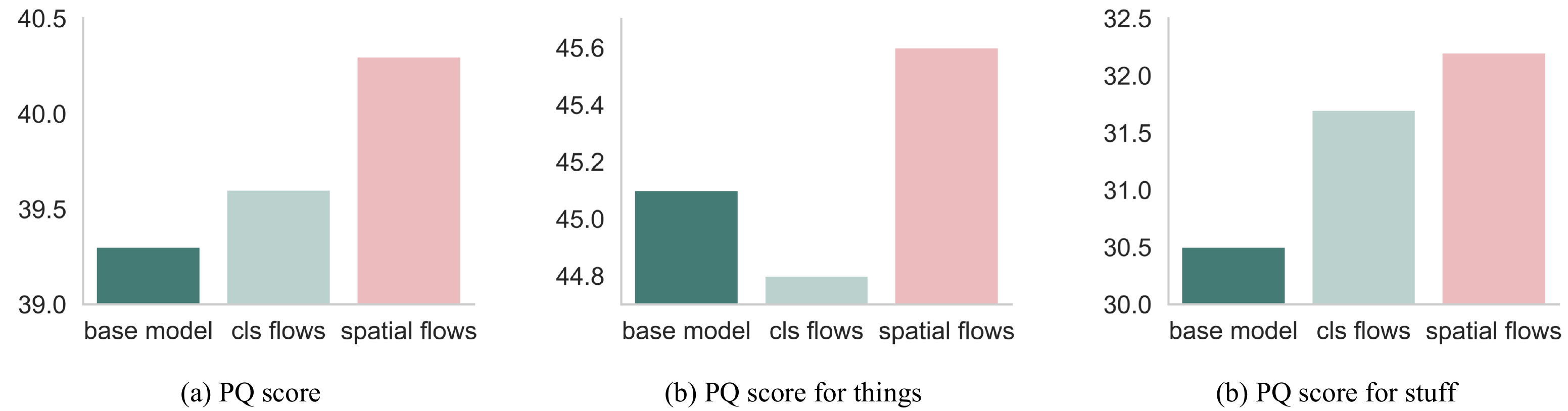}
  \caption{The PQ, PQ$^{Th}$, and PQ$^{St}$ results of  the base model, the cls flows model, and the spatial flows model with image size of 600px on COCO {\em val} split.}
   \label{fig3}
\end{figure*}
\\[3mm]
\textbf{Spatial Flows.} We conduct experiments to highlight the significance of the proposed spatial information flows between tasks. The baseline model marked with `-' in Table~\ref{tab8a} and Table~\ref{tab8b} is the one with all sub-networks. There are three paths to deliver the spatial context from the box regression task to others: the path from the reg sub-net to the cls sub-net (reg-cls flow), the path to the stuff sub-net (reg-stuff flow), and the path to the thing sub-net (reg-thing flow). The results are reported in Table~\ref{tab8a} and Table~\ref{tab8b}. At first, we add the reg-cls path, and we obtain a $0.4$ PQ$^{Th}$ improvement on COCO and a $1.5$ PQ$^{Th}$ gain on Cityscapes, which are brought by better detection results. Adding spatial context helps cls sub-net to extract discriminative features, which is essential for detection. Then we build a spatial path for stuff sub-net, as shown in the $4$th row of Table~\ref{tab8a} and Table~\ref{tab8b}, we earn a $0.6$ PQ$^{St}$ gain on COCO and a $0.8$ PQ$^{St}$ gain on Cityscapes compared with the former model, which indicates that the spatial context begets a positive effect on stuff segmentation. The reg-thing path and the semantic path also show their effectiveness on both things and stuff segmentation. Comparing with the original model, SpatialFlow can achieve a consistent gain in both thing and stuff segmentation. The results prove the significance of the spatial context in panoptic segmentation to some extent. It is worth noting that we only apply the element-wise sum operation to integrate the spatial context in this work. We believe further improvement could be achieved with a more deliberate design like attention modules. 

\begin{figure}
  \centering
  \includegraphics[width=0.43\textwidth]{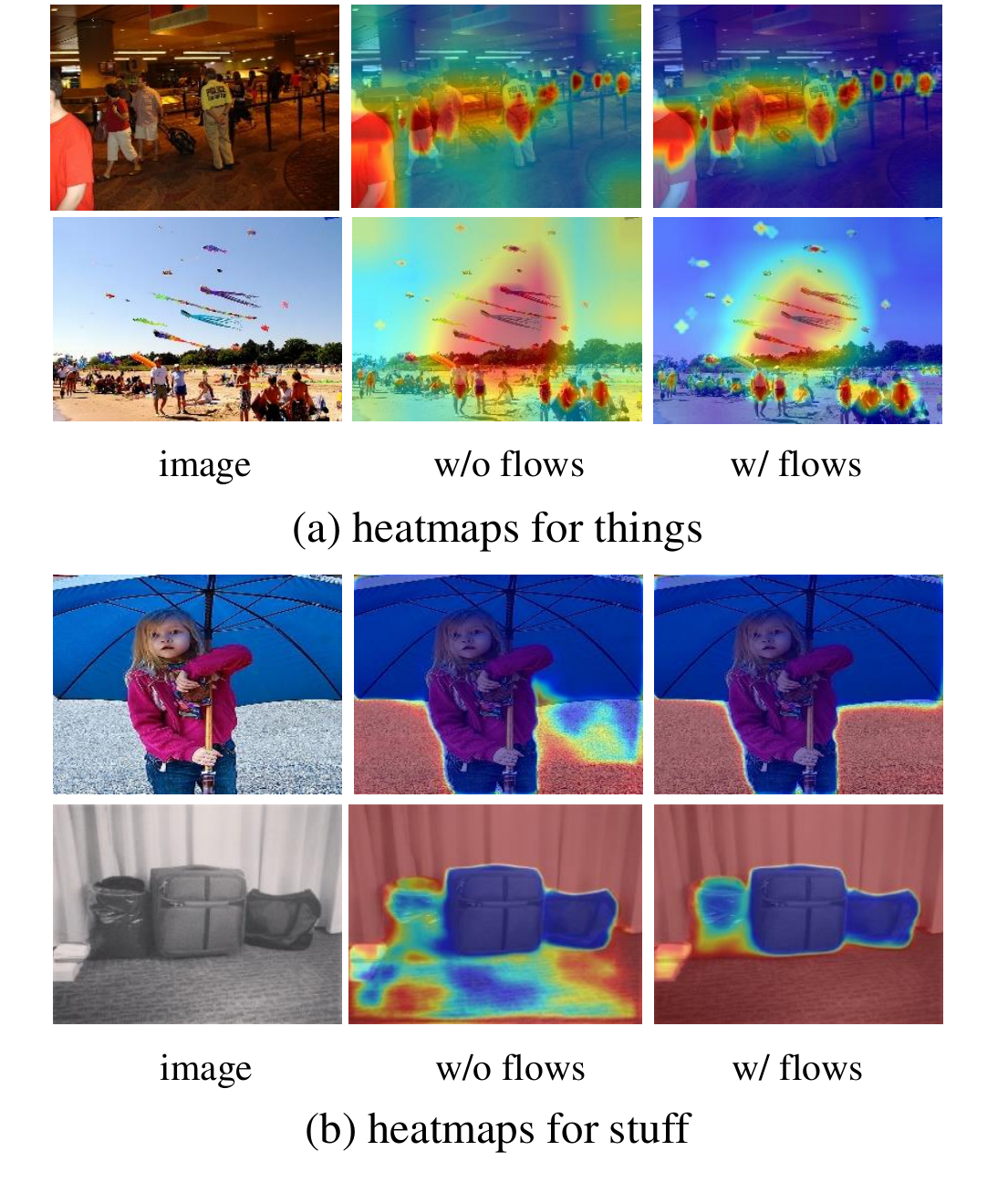}
  \caption{An illustration of the cls-head heatmap and the stuff-head heatmap. We provide a comparison between the model with and without spatial flows.}
   \label{fig4}
\end{figure}
\begin{figure*}
  \centering
  \includegraphics[width=1.0\textwidth]{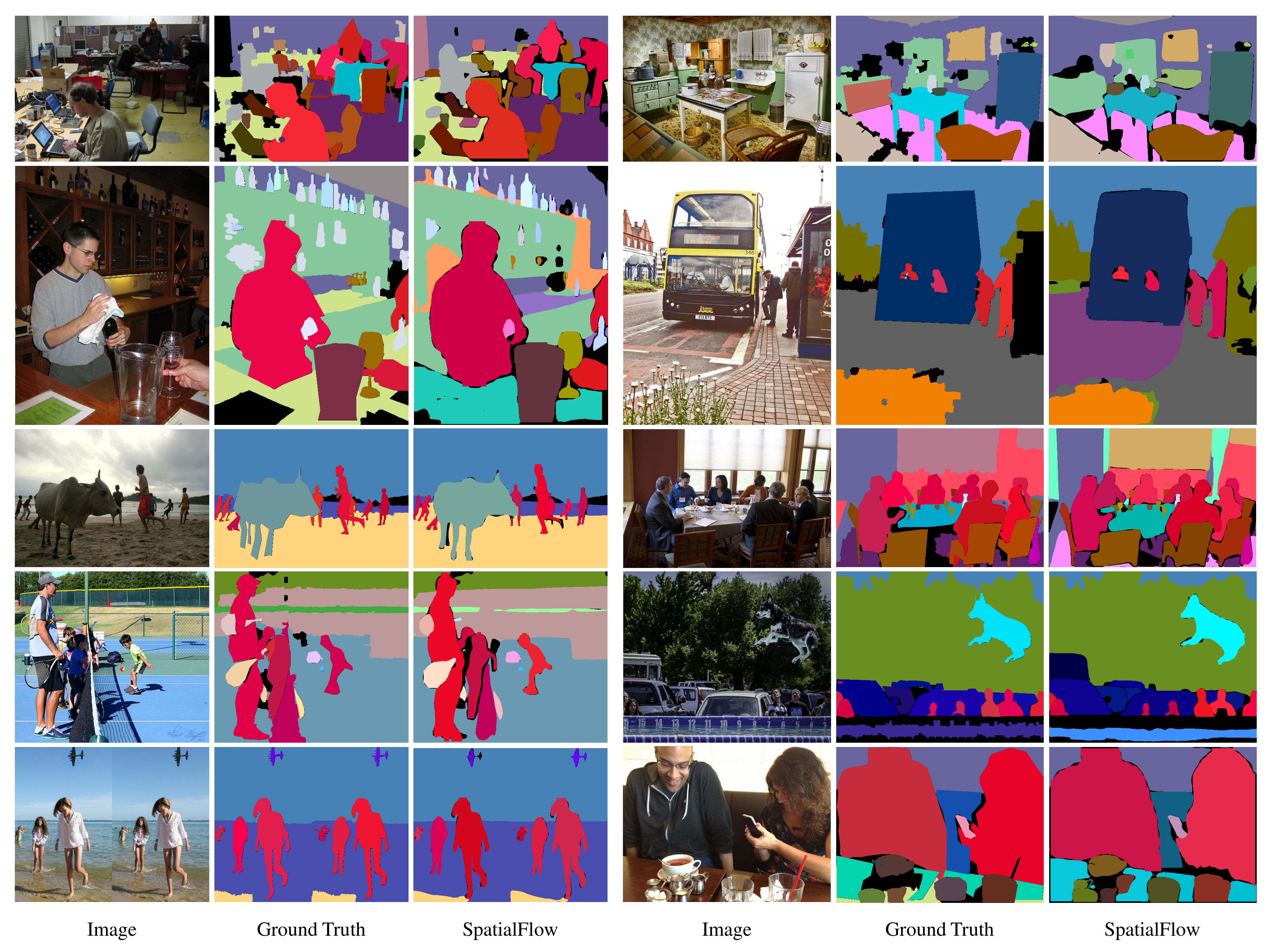}
   \caption{An illustration of visualization examples of SpatialFlow on COCO {\em val} split using a single ResNet-101 network.}
   \label{fig5}
\end{figure*}
\section{Further Discussion} \label{sec5}
In this section, we provide further discussions about the spatial flows and give an overview of how the spatial flows work, how fast the SpatialFlow is, and how to apply the spatial flows to other vision tasks.
\\[3mm]
\textbf{Spatial Flows {\em vs.} Trivial Feature Fusion:} Our main idea is to integrate spatial information into all sub-tasks and make them aware of the locations of the objects, which is different from trivial feature integration among sub-networks. To prove the effectiveness of the spatial flows, we design an experiment on COCO by delivering the feature of cls sub-networks to other three sub-tasks, denoted as the cls flows model in Figure~\ref{fig3}. We conduct an experiment on it with the image input size of 600px. As shown in Figure~\ref{fig3}, our method outperform the cls-based feature integration method by 0.7 PQ, 0.8 PQ$^{Th}$, and 0.5 PQ$^{St}$ respectively. The results suggest that trivial feature integration can not bring consistent improvements to the baseline model as our method does.
\\[3mm]
\textbf{Effects of Spatial Flows:} We choose to study the effects of spatial flows using two models, which are the models with or without spatial flows. We visualize the last feature map in the cls-head and the stuff-head of both models via CAM~\cite{cam} in Figure~\ref{fig4}. The visualized heatmaps illustrate that the spatial flows can help the thing branch focus on objects and make the stuff branch aware of the precise boundary of things and stuff. The spatial flows bridge all tasks and help build a global view of the image in panoptic segmentation.
\\[3mm]
\textbf{Accuracy {\em vs.} Speed:}  In Table~\ref{tab9}, we compare our method with the state-of-the-art methods in terms of accuracy and speed balance on COCO {\em val} split. The FPS is measured on a single Tesla V100 GPU. We show the results of different image sizes and different inference speed. Although SpatialFlow is not the fastest among all the methods, the results show good accuracy and speed balance of SpatialFlow. Larger image size yield higher accuracy, in slower inference speeds. Also, we find that thing segmentation benefits from large image sizes, while stuff segmentation is robust to the image size. Thanks to this, SpatialFlow can achieve $19.6$ FPS and remain $37.4$ PQ when we set the image size to $400$px.
\\[3mm]
\textbf{Detection Results:}  We also conduct experiments on RetinaNet~\cite{retinanet} to investigate the generalization of the spatial flows. We deliver the spatial context from reg sub-network to cls sub-network.  The detection result is shown in Table~\ref{tab10}. With the help of the spatial context, the multi-stage features in sub-networks can be more discriminative, which boosted the performance.
\begin{table}
  \centering
  \caption{\textbf{Accuracy {\em vs.} Speed}: Comparison with the state-of-the-art methods on accuracy and speed balance. We illustrate SpatialFlow performance with different image scales. * indicates that UPSNet apply deformable convolution on the stuff head.}
  \label{tab9}
  \begin{tabular}{ccccccc}
    \toprule
    Model & Backbone & Scale & PQ & PQ$^{Th}$ & PQ$^{St}$ & FPS\\
    \midrule
    \midrule
    PanopticFPN~\cite{panopticfpn} & ResNet-50 & 800 & 39.0 & 45.9 & 27.9 & 18.9 \\
    UPSNet*~\cite{upsnet} & ResNet-50 & 800 & 42.5 & 48.5 & 33.4 & 9.1 \\
    DeeperLab~\cite{deeperlab} & Xception-71 & 641 & 34.3 & 37.5 & 29.6 & 10.6 \\
    \midrule
    SpatialFlow & ResNet-50 & 800 & 40.9 & 46.8 & 31.9 & 10.3\\
    SpatialFlow & ResNet-50 & 600 & 40.3 & 45.6 & 32.2 & 13.0\\
    SpatialFlow & ResNet-50 & 400 & 37.4 & 41.5 & 31.4 & 19.6\\
    \bottomrule
  \end{tabular}
\end{table}
\begin{table}
  \centering
  \caption{\textbf{Detection Results}: The results of RetinaNet with or without spatial flows on COCO {\em val} split with ResNet-50 as the backbone. The shorted edges of images are 800px. }
  \label{tab10}
  \begin{tabular}{cccc}
    \toprule
    Detectors & mAP & AP$_{50}$ & AP$_{75}$ \\
    \midrule
    \midrule
    RetinaNet & 35.6 & 55.5 & 37.7 \\
    RetinaNet w/ flows & \textbf{36.7} & \textbf{57.1} & \textbf{39.4} \\
    \bottomrule
  \end{tabular}
\end{table}

\section{Visualization} \label{sec6}
\begin{figure*}
  \centering
  \includegraphics[width=1.0\textwidth]{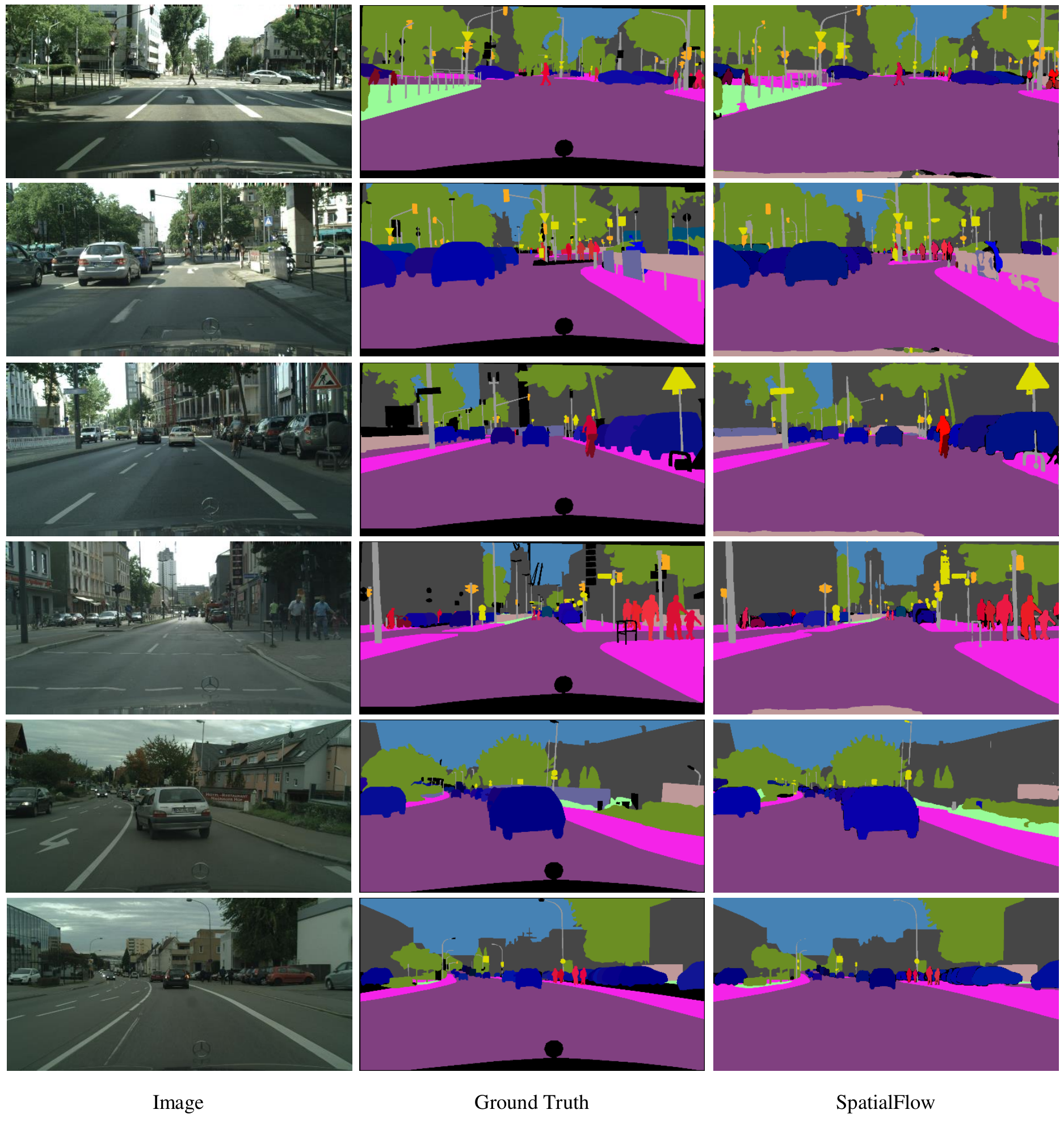}
  \caption{An illustration of visualization examples of SpatialFlow on Cityscapes {\em val} split using a single ResNet-101 network.}
   \label{fig6}
\end{figure*}

We show some visualization examples of SpatialFlow on COCO and Cityscapes in Figure~\ref{fig5} and Figure~\ref{fig6} respectively.

\section{Conclusion} \label{sec7}
In this work, we focus on the box locations in panoptic segmentation and propose a new location-aware and unified framework, denoted as SpatialFlow. We emphasize the importance of the spatial context and bridge all the tasks by building spatial information flows, then achieve state-of-the-art performance on both COCO {\em test-dev} split and Cityscapes {\em val} split, which prove the effectiveness of our model. Moreover, we find that the spatial flows can improve the performance of detection models, indicating the importance of spatial information. We expect that SpatialFlow can provide valuable insights on how to integrate spatial information in vision tasks.

\section*{Acknowledgment}
This work was supported in part by National Natural Science Foundation of China (No.61972396, 61876182, 61906193), National Key Research and Development Program of China (No. 2019AAA0103402),  the Strategic Priority Research Program of Chinese Academy of Science(No.XDB32050200), the Advance Research Program (No. 31511130301), and Jiangsu Frontier Technology Basic Research Project (No. BK20192004). Moreover, the authors would like to thank Jiaying Guo at Nanjing Institute of Geography and Limnology, Chinese Academy of Sciences for valuable discussions about the writing.

\ifCLASSOPTIONcaptionsoff
  \newpage
\fi



\bibliographystyle{IEEEtran}
\bibliography{IEEEbio}
\end{document}